\documentclass[10pt,twocolumn,letterpaper]{article}

\pdfoutput=1

\usepackage{cvpr}
\usepackage{times}
\usepackage{epsfig}
\usepackage{graphicx}
\usepackage{amsmath}
\usepackage{amssymb}

\usepackage{multirow}
\usepackage[ruled,vlined]{algorithm2e}
\usepackage{gensymb}
\usepackage{float}
\usepackage{balance}
\usepackage{natbib}

\setcitestyle{numbers,open={[},close={]}}

\usepackage[pagebackref=true,breaklinks=true,letterpaper=true,colorlinks,bookmarks=false]{hyperref}

\cvprfinalcopy 

\def\cvprPaperID{10275} 

\ifcvprfinal\fi
\begin{document}

\title{Laguerre-Gauss Preprocessing:\\ Line Profiles as Image Features for Aerial Images Classification}

\author{Alejandro Murillo-Gonz\'alez\\
Universidad EAFIT\\
{\tt\small amurillog@eafit.edu.co}
\and
Jos\'e David Ortega Pab\'on\\
Fuerza A\'erea Colombiana\\
{\tt\small jose.ortega@fac.mil.co}
\and
Juan Guillermo Paniagua\\
Universidad EAFIT\\
{\tt\small juanpaniagua@itm.edu.co}
\and
Olga Luc\'ia Quintero Montoya\\
Universidad EAFIT\\
{\tt\small oquinte1@eafit.edu.co}
}

\maketitle
\thispagestyle{empty}   

\begin{abstract}
    An image preprocessing methodology based on Fourier analysis
    together with the Laguerre-Gauss Spatial Filter is proposed.
    This is an alternative to obtain features from aerial images
    that reduces the feature space significantly, preserving enough
    information for classification tasks.
    Experiments on a challenging data set of aerial images show that it is possible to
    learn a robust classifier from this transformed and smaller feature
    space using simple models, with similar performance to
    the complete feature space and more complex models.
\end{abstract}

\section{Introduction}
Image preprocessing techniques aim to enhance an image's relevant features
and suppress those that are considered noise for the task at hand \cite{Sonka1993}.
This process involves transforming the representation, usually a matrix
filled with intensity values, following a predefined procedure. The
desire to obtain an enhanced or smaller representation comes from
different areas such as medical diagnosis, information transmission
and compression, digital photography and computer vision (CV).

An effective preprocessing might result in the solution of
CV problems with simple learning algorithms.
This simple models make a more efficient use of the data,
which means a time reduction during inference and training,
something that is crucial for real-time applications or problems
that require online learning. Also, the memory and disk space
footprint is reduced. It is important to note that simpler
solutions are easier to interpret most of the time, resulting
in models whose inner workings are clearer.

Therefore, this work explores the applicability of procedures
that reduce an image's feature space to perform classification
tasks using k-Nearest Neighbors (kNN) \cite{NearestNeighbor} 
and Multilayer Perceptrons (MLP) \cite{Rosenblatt58theperceptron,Goodfellow-et-al-2016},
which are models currently displaced -in this task- by more
powerful but expensive models such as Convolutional Neural
Networks (CNN) \cite{NIPS2012_4824, Goodfellow-et-al-2016}.

The employed approach is based on spectrum analysis tools,
particularly Fourier analysis. It uses the Laguerre-Gauss 
Spatial Filter (LGSF), as kernel, proposed
by Gou \etal \cite{Guo:06}, who state that ``it allows the
realization of a radial Hilbert transform with high contrastive
and isotropic edge enhancement without resolution loss". 
This means that the borders will be intensified (in all directions),
thus making them easier to detect. In \cite{RBGf822} the authors
compare LGSF against other common edge enhancing
technique (Laplacian filtering) and found it to be superior
given that it acts as a bandpass reducing low and high
frequency noise, and that it ``distributed homogeneously and 
smoothly the intensity of the magnitude of the Fourier
spectrum due to its isotropic feature". The latter statements
are an indication that a line profile, along an axis of an image
preprocesed with this filter, will result in a coherent curve. 
Is this line profile, taken along the x- and y-axis, that will
be used in this work as the feature space during classification. Furthermore,
the results from Sierra-Sosa \etal \cite{Sierra_Sosa_2016} showed
that, after Double Fourier analysis, the line profile of a
voiced speech was useful to classify emotions.

Using the feature space obtained from the principles described
above, it was possible to obtain good performance on multiple
classification tasks such as Geometric Shapes and Aerial Images,
where CNN were replaced by kNN and MLP. As a preliminary result,
this preprocessing methodology could be extended to other
CV tasks.

Concretely, the contribution of this paper is the description
of an image preprocessing methodology to reduce the dimensionality
of the data and perform classification with simple models. Particularly, a challenging data set of Aerial Images will be classified using this method.

The code implementation of Laguerre-Gauss Preprocessing and the
presented results is available on \href{https://github.com/AlejandroMllo/Laguerre-Gauss-Preprocessing}{GitHub} \footnote{\url{https://github.com/AlejandroMllo/Laguerre-Gauss-Preprocessing}}.

The remaining of this article is organized as follows. Section
\ref{sec:sota} gives an overview of the related work. Section
\ref{sec:preliminaries} goes over the preliminary theoretical
framework. In Section \ref{sec:methodology} the proposed methodology
for image feature space reduction and classification is described.
Section \ref{sec:results} presents the data sets employed and the
results obtained. Section \ref{sec:discussion} discusses the
results. Finally, Section \ref{sec:conclusion} introduces the
reached conclusions.

\section{Related Work}
\label{sec:sota}

\textbf{Line Profiles.} Sierra-Sosa \etal \cite{Sierra_Sosa_2016} use Fourier analysis on spectrograms of audio 
recordings to identify the speaker's emotions using the spectrum's line profile. In \cite{logo_lineprofile} multiple line profiles, whose position are based on maxima in the 
Hough transform space, are used to recognize logos (classification over invariant classes). In \cite{lineprofile_eyereflections} the authors employ line intensity profiles to identify reflections in 
eye images, which are then classified as reflections or non-reflections using a suport vector 
machine. Bhan \etal \cite{lineprofile_caries} use feature line profiles from preprocessed radiographic images to detect dental caries and their severity.

The works described above, specially those on images, work over very restricted spaces. In the Aerial Images classification
task the environments and elements of each class have higher variation.

\textbf{Aerial Images.} In \cite{cnn_objectdetection, Radovic_2017} the authors propose CNNs to detect objects from aerial images. Qayyum \etal \cite{sceneclass_aerial} use CNN to classify scenes in aerial images extracting features in multiple scales which are then encoded into global image features.

\section{Preliminaries}
\label{sec:preliminaries}

This section introduces the definitions and theorems (based on \cite{debnath2006integral}) behind Laguerre-Gauss Preprocessing, which is described in Section \ref{sec:methodology}.

\subsection{Integral Transforms} \label{sec:int_trans}
\noindent \textbf{Def.} An \textit{Integral Transform} is defined as
\begin{equation*} \label{int_trans} \tag{1}
\mathcal{I}\{f(\textbf{x})\} = F(\textbf{k}) = \int_s \mathcal{K}(\textbf{x}, \textbf{k})f(\textbf{x}) d\textbf{x}
\end{equation*}

where $\textbf{x} = (x_1, x_2, \dots, x_n), \textbf{k} = (k_1, k_2, \dots, k_n), S \subset \mathbb{R}^n$ and $\mathcal{K}$ is the kernel of the transform. The \textit{inverse transform} is an operator such that

\begin{equation*} \tag{2} \label{inv_int_trans}
    \mathcal{I}^{-1}\{F(\textbf{k})\} = f(\textbf{x})
\end{equation*}{}

\subsubsection{Convolution}

\textbf{Def.} The \textit{Convolution} of two integrable functions $f(\textbf{x})$ and $g(\textbf{x})$, denoted by $(f \ast g)(\textbf{x})$, is
\begin{equation*} \tag{3}
    (f \ast g)(\textbf{x}) = \int_{-\infty}^{\infty} f(\textbf{x} - \xi)g(\textbf{x})d\xi
\end{equation*}

\noindent \textbf{Convolution Theorem.} If $\mathcal{I}\{f(\textbf{x})\} = F(\textbf{k})$ and $\mathcal{I}\{g(\textbf{x})\} = G(\textbf{k})$, then
\begin{align*}
    \mathcal{I}\{f(\textbf{x}) \ast g(\textbf{x})\} & = F(\textbf{k})G(\textbf{k}) \\ \tag{4}
    f(\textbf{x}) \ast g(\textbf{x}) & = \mathcal{I}^{-1}\{F(\textbf{k})G(\textbf{k})\}
\end{align*}

Furthermore, commutativity, associativity and distributivity is valid under a convolution.

\subsubsection{Fourier Transform}

A particular case of an integral transform is the Fourier transform which turns a function's (signal) time domain into its frequency representation.

\noindent \textbf{Def.} The \textit{Multiple Fourier Transform} of $f(\textbf{x})$, where $\textbf{x} = (x_1, x_2, \dots, x_n)$ is the \textit{n}-dimensional vector, is defined by
\begin{equation*} \tag{5}
    \mathcal{F}\{f(\textbf{x})\} = \frac{1}{(2\pi)^\frac{n}{2}}\int_{-\infty}^{\infty} \dots \int_{-\infty}^{\infty}\exp{\{-\textit{i}(\textbf{k}\cdot\textbf{x})\}}f(\textbf{x})d\textbf{x}
\end{equation*}{}
where $\textbf{k} = (k_1, k_2, \dots, k_n)$ is the 
\textit{n}-dimensional vector, $\textbf{k}\cdot\textbf{x} = k_1x_1 + 
k_2x_2 + \dots + k_nx_n$ and $f(\textbf{x})$ is absolutely integrable.

\subsection{Hilbert Space} \label{sec:hilbert_space}

\noindent \textbf{Def.} Let $\mathcal{H}$ be a vector space. A bi-linear mapping $<\cdot, \cdot> : \mathcal{H} \times \mathcal{H} \xrightarrow{} \mathbb{R}$ is an \textit{inner product} if it satisfies symmetry, 
linearity and positive definiteness.

\noindent \textbf{Def.} A vector space $\mathcal{H}$ is a \textit{Hilbert space}
if it is a complete inner product space.

\noindent \textbf{Def.} Given a Hilbert space $\mathcal{H}$, a
\textit{feature mapping} $\phi : x \xrightarrow{} \mathcal{H}$ is a
function such that $\phi(x) \in \mathcal{H}$.

\noindent \textbf{Theorem.} Any feature mapping defines a valid kernel.

\vspace{5mm}

Note that having a Hilbert space enables transformations in any number 
of dimensions. Once a kernel is obtained, a transformation, as defined in \ref{int_trans},
can be performed using it.

\subsection{Laguerre-Gauss Spatial Filter (LGSF)} \label{sec:lgsf}

The LGSF is the kernel employed during preprocessing to enhance the edges
and reduce low- and high- frequency noise \cite{Guo:06, RBGf822}.

The filter, proposed by Guo \etal \cite{Guo:06}, in the spatial domain is
given by \cite{attenuation_paniagua}

\begin{equation*} \tag{6}  \label{spatial_lg}
    LG(x, y) = (\textit{i}\pi^2 \omega^4)(x + \textit{i}y)\exp{\{-\pi^2\omega^2(x^2 + y^2)\}}
\end{equation*}{}

where $\omega$ is a parameter that controls the bandpass filter size. As $\omega$ approaches 1 it favors higher frequencies, that is, thinner edges.

\subsection{Line Profile}
The line profile of a matrix is the sampling of its values along a path.

This work uses the line profiles taken along a line segment that crosses
the origin and is parallel to the $x-$ and $y-$axis.

In the case of images in the spatial domain, the line profile samples the
intensity value of its pixels. If the image is in the frequential domain,
the line profile samples a frequency, which is a complex number $z$. To
visualize the sampled frequencies it is possible to take the amplitude
(which is the absolute value of $z$), its phase, its real or its imaginary
component.

\section{Proposed Method}
\label{sec:methodology}

\begin{figure}[H]
\begin{center}
\includegraphics[width=\linewidth]{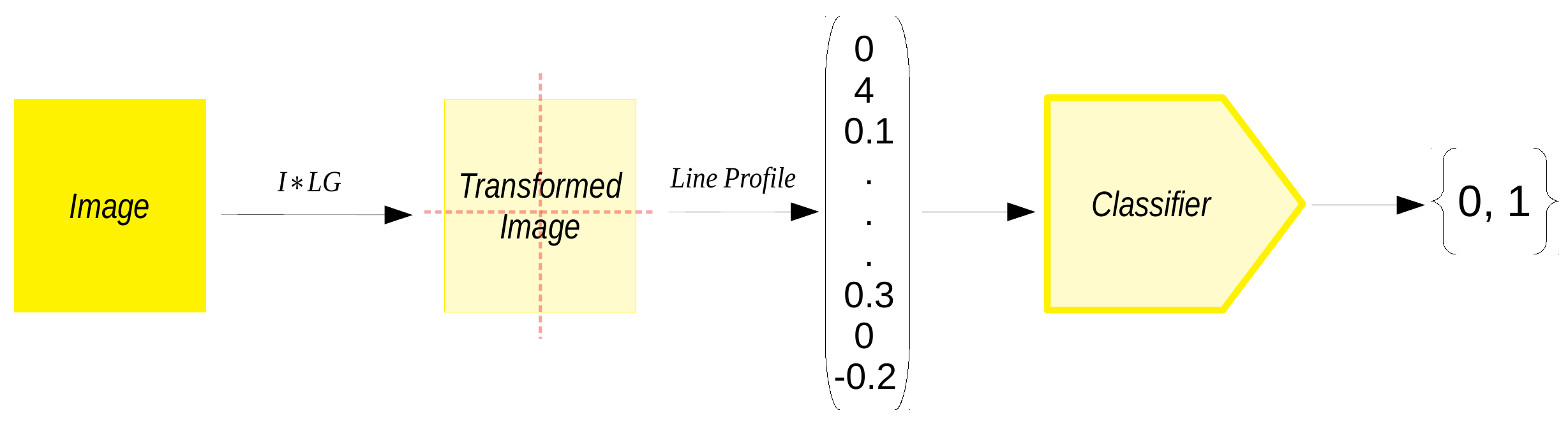}
\end{center}
   \caption{Overview of Laguerre-Gauss Preprocessing.}
\label{fig:LGPrep_overview}
\end{figure}

\subsection{Overview}

Figure \ref{fig:LGPrep_overview} summarizes the Laguerre-Gauss 
preprocessing procedure. The methodology is based in the 
application of integral transforms as described in 
Section \ref{sec:int_trans}. It acts on a function to produce another 
simpler representation. This might also happen to be interpretable 
and enable comparison between other functions who undergone the same 
transformation process. 

Particularly, this transformation process involves
getting the Fourier's spectrum representation from the image $\hat{I}$ and
the LGSF $\hat{LG}$ filter introduced in Section \ref{sec:lgsf}. Once both analytic
signals are obtained, given that they are no longer in the spatial
domain, their convolution is their point-wise multiplication $I \ast LG = \hat{I} \cdot \hat{LG}$. Here $LG$ is acting as a feature mapping, therefore it is
considered a valid kernel, as seen in Section \ref{sec:hilbert_space}.

The remaining steps involve shifting the zero-frequency components to the
center of the image to obtain an image whose origin is at its center. Then
the line profiles are sampled along the $x-$ and $y-$axis.

This procedure means that if an image of size $n \times n$ is being classified,
the classifier will only need $2 \times n$ features, reducing significantly the
feature space.


\subsection{Laguerre-Gauss Preprocessing}

In order to reduce the model's dimensionality, the input images will be preprocessed to obtain a set of features representative of the elements present on it.

Preprocesssing is as follows:
\begin{enumerate}
    \item Create the Laguerre-Gauss filter \cite{attenuation_paniagua} from equation \ref{spatial_lg} with the same size as the image and a given \(\omega\). It is usually set to \(\omega = 0.9\) to better distinguish the image's borders.
    \item Apply Fourier's transform to the filter and the image.
    \item Once the analytic signal of both the filter and the image is obtained, they are convolved. This is done through element-wise multiplication. Since the filter is 2-dimensional, the images should be transformed to one channel (i.e., grayscale) or apply the filter to each or a set of channels on a channel-wise basis.
    \item Next, the zero-frequency components are shifted to the center of the spectrum. This means that
    the origin of the image will be its center.
    \item Finally, the \(x\) and \(y\) line-profiles are obtained from the shifted image. This is
    the vector representing the spectrum along the origin in the \(x-\) and \(y-\) axis respectively.
\end{enumerate}

Algorithm \ref{lg_prep_algorithm} presents the pseudocode of the Laguerre-Gauss Preprocessing procedure.

\begin{algorithm}[] \label{lg_prep_algorithm}
\SetAlgoLined{}
\KwData{image, $\omega$}
s $\gets$ \texttt{size}(image); \\
filter $\gets$ \texttt{LaguerreGaussFilter}($\omega$, s); \\
$\text{image}_{FT}$ $~\gets$ \texttt{FourierTransform}(image); \\
$\text{filter}_{FT}$ $~~~\gets$ \texttt{FourierTransform}(filter); \\
convolved $\gets$ $\text{image}_{FT}$ $\cdot$ $\text{filter}_{FT}$; \\
shifted $\gets \texttt{shift}$(convolved); \\
\textit{x}-profile $\gets \texttt{LineProfile}(shifted, axis=x)$; \\
\textit{y}-profile $\gets \texttt{LineProfile}(shifted, axis=y)$; \\
\Return \textit{x}-profile, \textit{y}-profile
\caption{Laguerre-Gauss Preprocessing}
\end{algorithm}


\section{Results}
\label{sec:results}

This section describes the data set and classifiers employed, together
with the results from using each of them with Laguerre-Gauss preproprocessing,
and without it. Also, it presents the results from a couple of ablation
experiments to show the utility of this preprocessing methodology
as a whole.

\subsection{Data Sets}
\subsubsection{Geometric Shapes}

This data set is a mix of \cite{data_shapes_mark, data_shapes_herz, data_geometricshapes_rushan, data_fourshapes}
which combines computer generated and hand-drawn
geometric shapes. All images are resized to $64 \times 64$ pixels. Basically, the images are
monochromatic with different sizes, rotations and deformations. Specifically, the shapes
to be classified are Circles (which includes ellipses), Squares (which includes rectangles)
and Triangles. Table \ref{table:geom_shapes_dataset} shows the data set specificities.

\begin{table}[H]
\begin{center}{}
\begin{tabular}{|c|ccc|c|}
\hline
\textbf{Set} & \multicolumn{1}{c}{\textbf{Circles}} & \multicolumn{1}{c}{\textbf{Squares}} & \textbf{Triangles} & \textbf{Total} \\ \hline
Train        & 3277       & 4058            & 3608       & 10943          \\ 
Validation   & 859          & 803             & 720        & 2382           \\ 
Test         & 889           & 847         & 600         & 2336           \\ \hline
\end{tabular}
\caption{Description of Geometric Shapes data set.}
\label{table:geom_shapes_dataset}
\end{center}
\end{table}

\subsubsection{Aerial Images}

This data set was collected to identify illegal mining and deforestation in large areas.
The images come from multiple news agencies \cite{mineria_caracol, mineria_rcn, mineria_semana, mineria_eltiempo}
which have covered this phenomena.
The idea is to classify an image in the class 1 if it contains something of interest
(i.e., heavy-equipment, boats, deforestation, etc.); or as class 0 if it does not (i.e., forest,
rivers, populations, etc.). The data set contains 5707 instances (resized to $64 \times 64$ pixels) of which 3297 (57\% of the 
total) images are samples of class 0 and the remaining are samples from class 1. The data
was augmented to 20000 instances through random rotations ($\pm 20\degree$, 90\degree, 
270\degree), left-to-right and top-to-bottom flips. 

%
%
%
%
%

The main challenge posed by this data set is the infinite number of possible camera angles and
altitudes from which the images are taken. Also, the noise added by images from moving
cameras, fog, clouds and changing climates. Note that not all images come from the same
distribution, since they are taken in different locations with different kinds of sensors.

\subsection{Classification}

Three kinds of learners (kNN, MLP \& CNN) were used to test the preprocessing 
methodology against no-preprocessing, or a flattened representation
of an image. 

The \textbf{kNN} employed a value of \(k = 1\) for the number
of neighbors in all tasks. 

The \textbf{MLP} used the \textit{same} architecture for all
tasks (regardless of the data representation), therefore the results reported
here might be improved with custom-made architectures. The MLP featured an
input layer with 128 units, two hidden layers with 64 and 32 units respectively,
and an output layer with as many units as the number of classes in the problem.
For regularization, a dropout layer was embedded after each of the three initial
layers, with probability 0.5, 0.25 and 0.25, respectively.

The \textbf{CNN} also used the \textit{same} architecture for all tasks. The 
architecture is shown in Figure \ref{fig:cnn_arch}. It was chosen from 
\cite{chollet2015keras}, such that it had an excellent performance on MNIST,
therefore the results for other data sets might be improved with different 
architectures.

Finally, both the MLP and CNN used categorical crossentropy as loss and RMSProp
as optimizer. Also, all their layers used relu as activation function, except the
last one which used softmax. No hyper-parameter tuning took place.

\subsubsection{Geometric Shapes}

Table \ref{table:geom_shapes_results} presents the classification results using 
three kinds of models (kNN, MLP and CNN). It describes the scores obtained using a
flattened representation of the image's pixel intensities matrix and compares it
with the classification scores using Laguerre-Gauss Preprocessing (LP).
A quick glance returns that the classifiers using LP did better, except in the case of
CNN, but considering the disk space trade-off and the inference time, the MLP might be
a better option for real-time and embedded applications, since the score's difference
is minimal.

\begin{figure*}[t]
\begin{center}
\includegraphics[width=\linewidth]{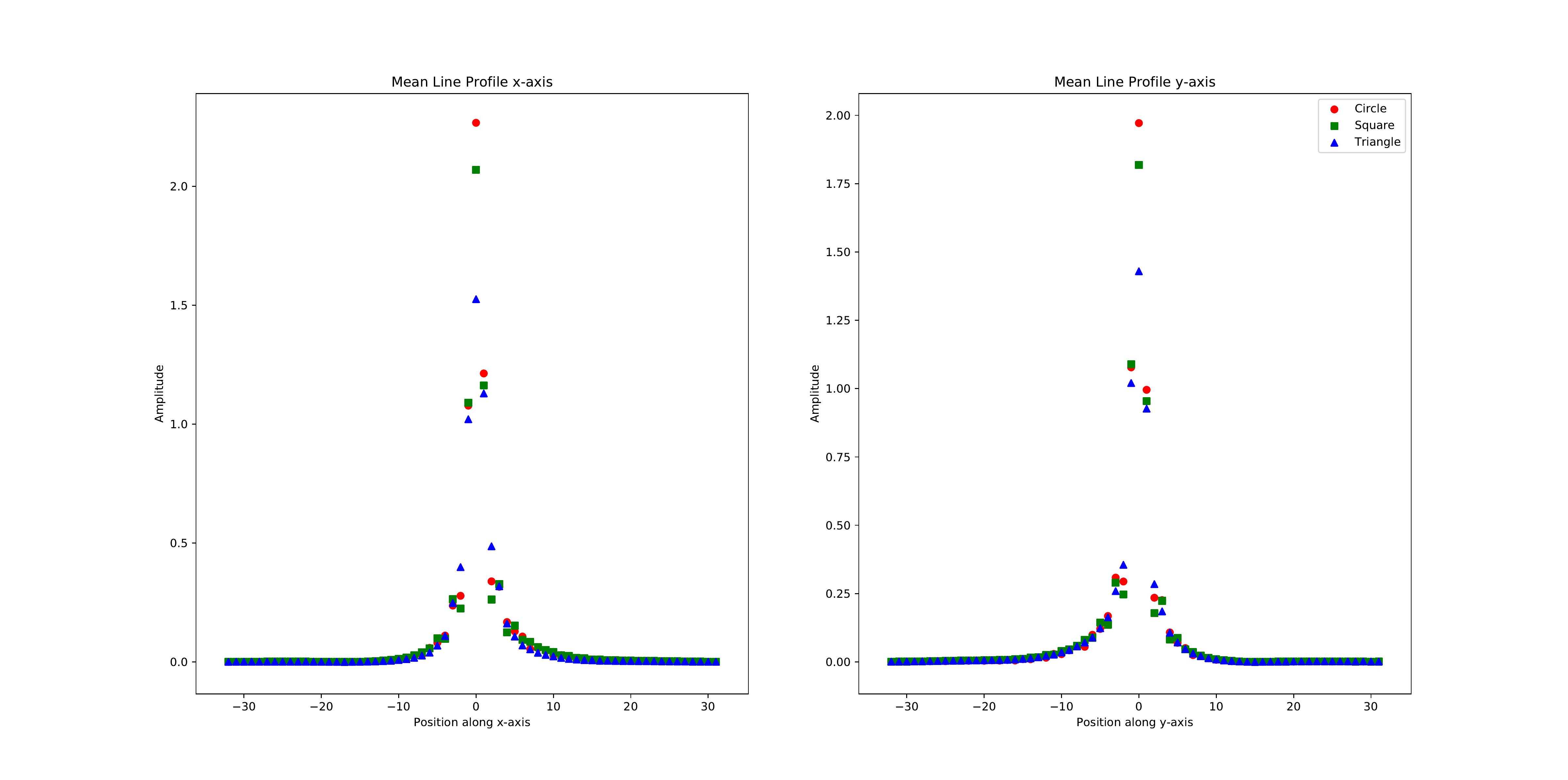}
\end{center}
   \caption{Mean Line Profile Geometric Shapes.}
\label{fig:lp_shapes}
\end{figure*}

Figure \ref{fig:lp_shapes} shows the mean line profiles for the train instances of the
geometric shapes. From it is possible to see the differences in the frequency spectrum
within the different shapes; this differences can also be appreciated by the learners.

\subsubsection{Aerial Images}

The results of the classification of aerial images are reported on Table 
\ref{table:aerial_images_results}. There it is possible to see that the MLP using the flattened
representation of the image cannot break the class imbalance as the accuracy never goes
above 57\%, which is confirmed by the F1 score. On the other hand, the MLP using the LP
representation does really well on the data set obtaining around 80\% accuracy. It is
remarkable the size difference of this model with the other ones tested. In the case
of the kNN implementation, even though the flattened representation has slightly better
performance, its size might be a constraint when deploying on resource constrained
environments. The huge difference in size is due to the fact that kNN need to store
the support seen during training, as a result a smaller feature space, such as LP, will
inevitably result in a smaller and faster model.

\begin{figure*}[t]
\begin{center}
\includegraphics[width=\linewidth]{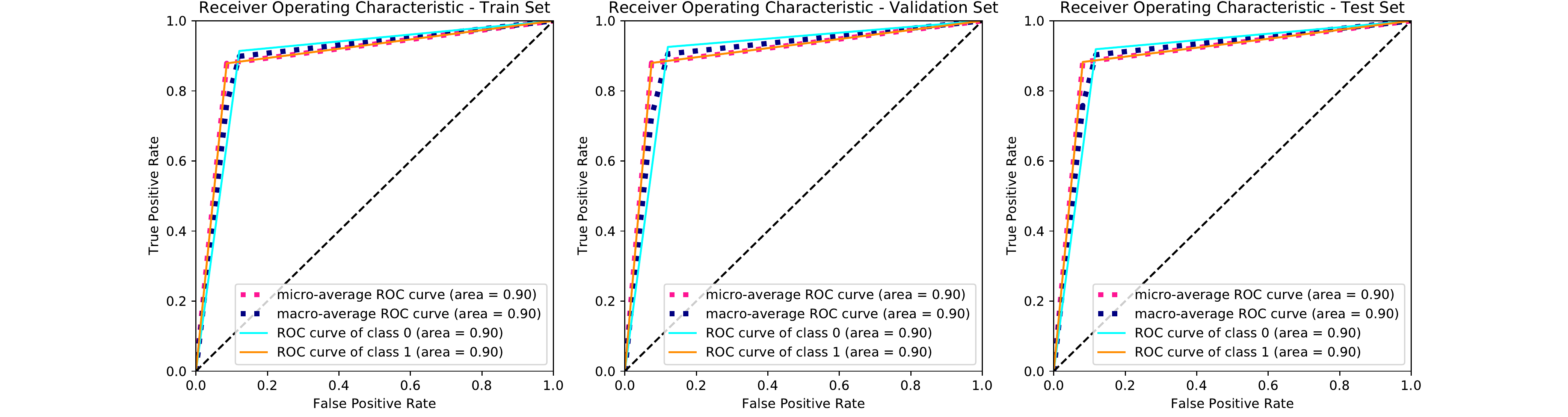}
\end{center}
   \caption{ROC curve for a kNN trained to classify Aerial Images.}
\label{fig:roc_aerial_knn}
\end{figure*}

\begin{figure*}[t]
\begin{center}
\includegraphics[width=\linewidth]{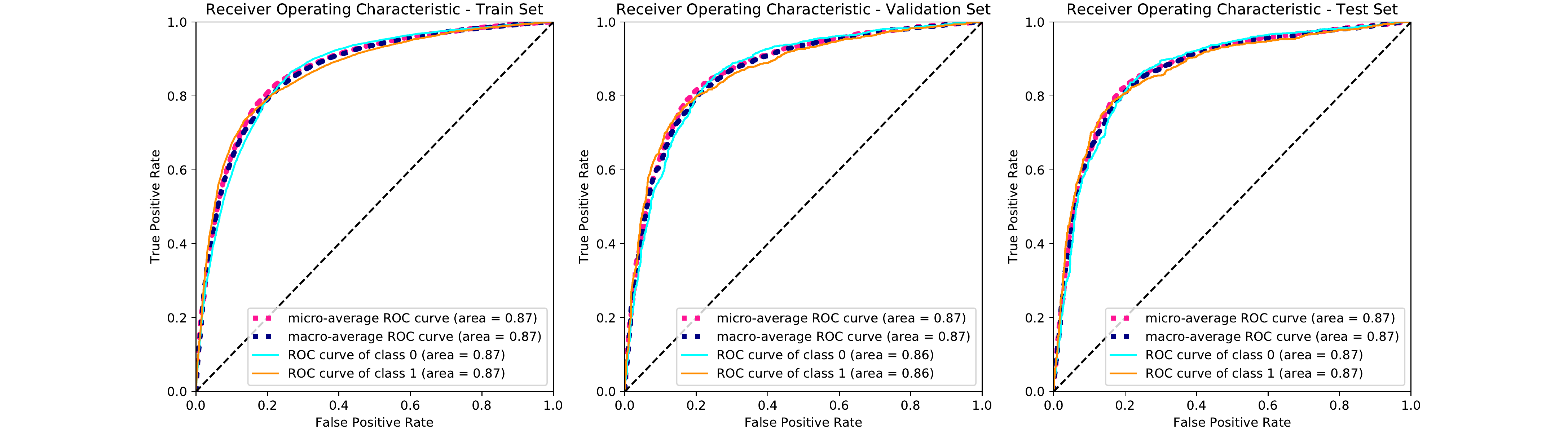}
\end{center}
   \caption{ROC curve for a MLP trained to classify Aerial Images.}
\label{fig:roc_aerial_mlp}
\end{figure*}

Figures \ref{fig:roc_aerial_knn} and \ref{fig:roc_aerial_mlp} show the Receiver
Operating Characteristic (ROC) curve of the kNN and MLP models employing the LP
feature space. The results displayed on the figures are a good indication of the combined
precision and recall of the models.

\begin{figure*}[t]
\begin{center}
\includegraphics[width=\linewidth]{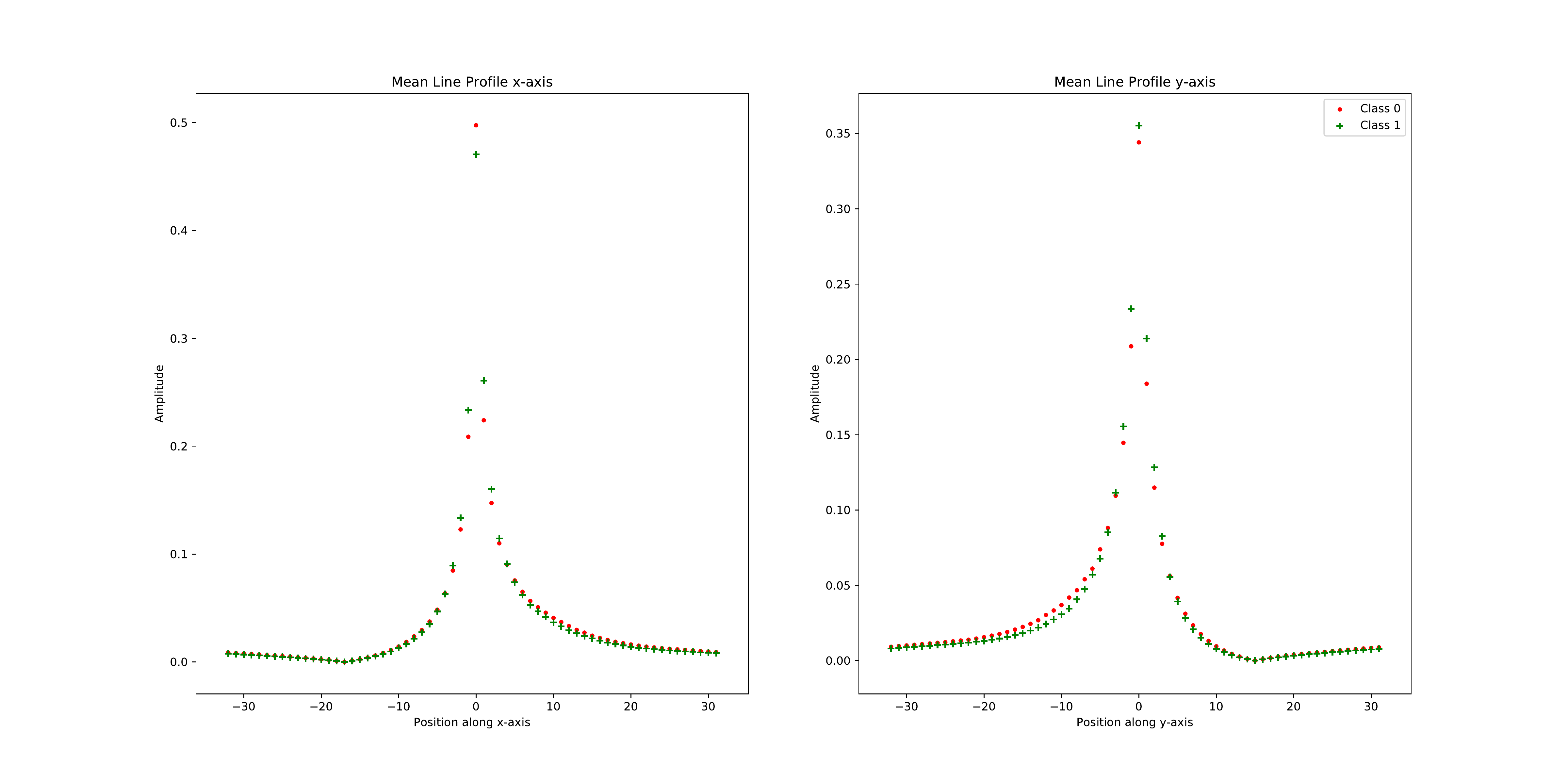}
\end{center}
   \caption{Mean Line Profile Aerial Images.}
\label{fig:lp_aerial}
\end{figure*}

Figure \ref{fig:lp_aerial} shows the mean line profiles for the classes in the data set. Here,
similar to the case of geometric shapes, the differences in the frequency spectrum are
observable, allowing the models to learn them too.

\begin{figure}[H]
\begin{center}
\includegraphics[width=0.5\linewidth]{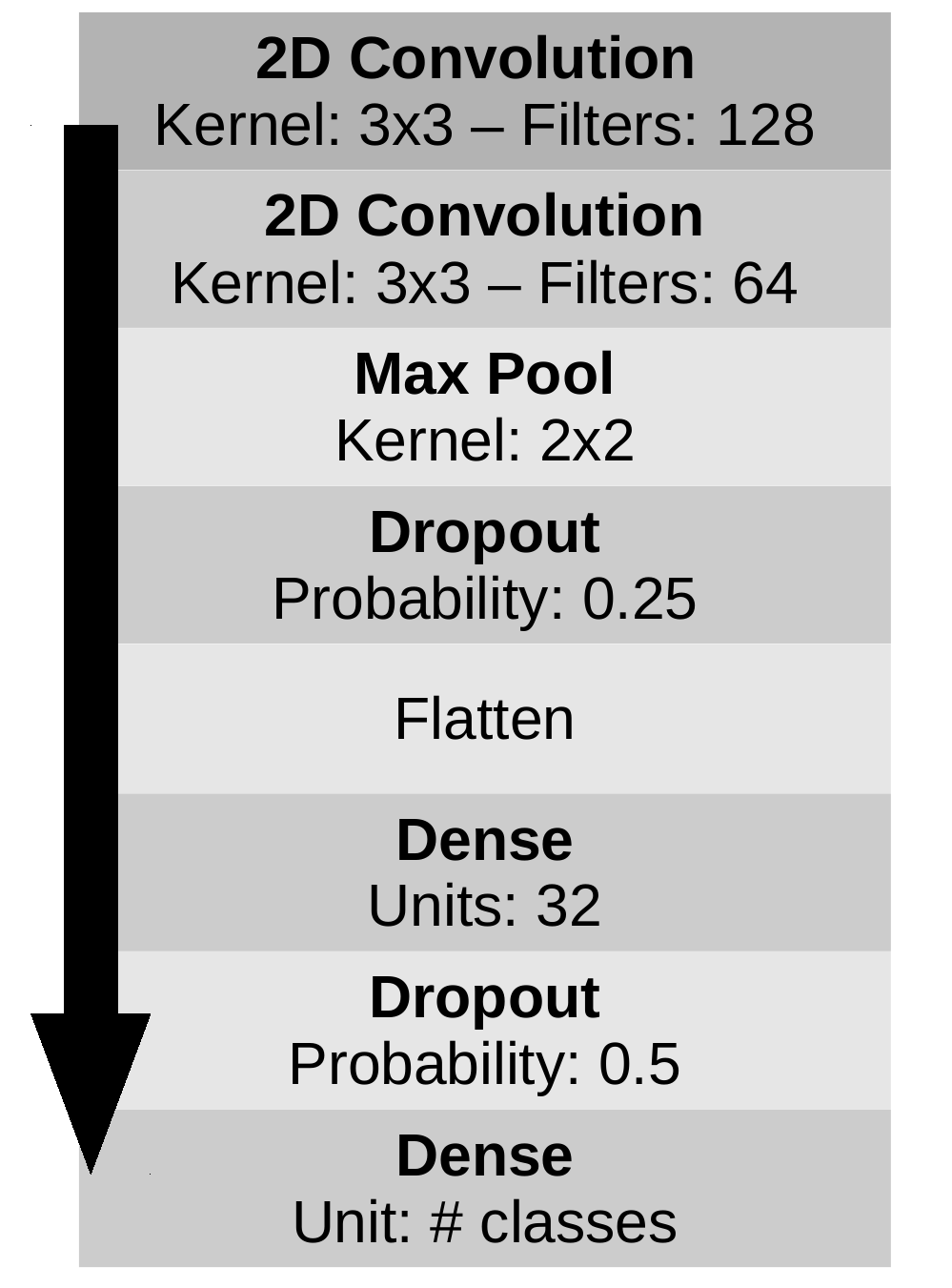}
\end{center}
   \caption{CNN Architecture employed for the classification tasks.}
\label{fig:cnn_arch_caption}
\label{fig:cnn_arch}
\end{figure}

The CNN from Figure \ref{fig:cnn_arch} cannot break the class imbalance, therefore
its results on the task are not reported.

\begin{table*}[] 
{\small
\begin{center}
\begin{tabular}{ccc|cc|cc|cc|}
\cline{4-9}
\textbf{}                                  & \textbf{}                          &                   & \multicolumn{2}{c|}{\textbf{Train}}  & \multicolumn{2}{c|}{\textbf{Validation}} & \multicolumn{2}{c|}{\textbf{Test}}   \\ \hline
\multicolumn{1}{|c|}{\textbf{Model}}       & \multicolumn{1}{c|}{\textbf{Data}} & \textbf{Size}     & \textbf{Accuracy} & \textbf{F1}      & \textbf{Accuracy}   & \textbf{F1}        & \textbf{Accuracy} & \textbf{F1}      \\ \hline
\multicolumn{1}{|c|}{\multirow{2}{*}{kNN}} & \multicolumn{1}{c|}{Flattened}     & 750.8 MB          & -                 & -                & 0.9513              & 0.94/0.94/0.96     & 0.9601            & 0.95/0.95/0.97   \\
\multicolumn{1}{|c|}{}                     & \multicolumn{1}{c|}{LP}            & \textbf{23.7 MB}  & -                 & -                & \textbf{0.9832}     & 0.97/0.98/0.99     & \textbf{0.9880}   & 0.98/0.98/0.99   \\ \hline
\multicolumn{1}{|c|}{\multirow{2}{*}{MLP}} & \multicolumn{1}{c|}{Flattened}     & 4.4 MB            & 0.9661            & 0.95/0.95/0.98 & 0.9311              & 0.93/0.90/0.95   & 0.9066            & 0.90/0.88/0.94 \\
\multicolumn{1}{|c|}{}                     & \multicolumn{1}{c|}{LP}            & \textbf{374.7 kB} & \textbf{0.9770}   & 0.96/0.97/0.98   & \textbf{0.9563}     & 0.94/0.95/0.97     & \textbf{0.9563}   & 0.95/0.95/0.96   \\ \hline
\multicolumn{1}{|c|}{CNN}                  & \multicolumn{1}{c|}{Image}         & 15.5 MB           & \textbf{0.9819}   & 0.98/0.97/0.98   & 0.9550              & 0.95/0.93/0.97     & 0.9584            & 0.96/0.94/0.96   \\ \hline
\end{tabular}
\end{center}{}
\caption{Results of Geometric Shape classification. The shapes are a combination of hand-drawn and computer generated with different sizes. The F1 score is reported for each class in the next order: Circle, Square, Triangle.}
\label{table:geom_shapes_results}
}
\end{table*}

\begin{table*}[]
\begin{center}
\begin{tabular}{ccc|cc|cc|cc|}
\cline{4-9}
\textbf{}                                  & \textbf{}                          &                   & \multicolumn{2}{c|}{\textbf{Train}} & \multicolumn{2}{c|}{\textbf{Validation}} & \multicolumn{2}{c|}{\textbf{Test}} \\ \hline
\multicolumn{1}{|c|}{\textbf{Model}}       & \multicolumn{1}{c|}{\textbf{Data}} & \textbf{Size}     & \textbf{Accuracy}   & \textbf{F1}   & \textbf{Accuracy}      & \textbf{F1}     & \textbf{Accuracy}   & \textbf{F1}  \\ \hline
\multicolumn{1}{|c|}{\multirow{2}{*}{kNN}} & \multicolumn{1}{c|}{Flattened}     & 951.2 MB          & \textbf{0.9257}     & 0.93/0.91     & \textbf{0.9286}        & 0.93/0.91       & \textbf{0.9183}     & 0.92/0.90    \\
\multicolumn{1}{|c|}{}                     & \multicolumn{1}{c|}{LP}            & \textbf{30.0 MB}  & 0.9030              & 0.91/0.88     & 0.8900                 & 0.90/0.86       & 0.9046              & 0.9151/0.89  \\ \hline
\multicolumn{1}{|c|}{\multirow{2}{*}{MLP}} & \multicolumn{1}{c|}{Flattened}     & 4.4 MB            & 0.5747              & 0.72/0.0        & 0.5720                 & 0.72/0.0          & 0.5730              & 0.72/0.0       \\
\multicolumn{1}{|c|}{}                     & \multicolumn{1}{c|}{LP}            & \textbf{376.4 kB} & \textbf{0.8012}     & 0.82/0.76     & \textbf{0.8116}        & 0.83/0.77       & \textbf{0.7990}     & 0.82/0.76    \\ \hline
\end{tabular}
\end{center}
\caption{Results of Aerial Images classification. The F1 score is reported for each class in the next order: 0 - No object of interest / 1 - object of interest.}
\label{table:aerial_images_results}
\end{table*}

\begin{table*}[]
\begin{center}
\begin{tabular}{cc|cc|cc|cc|}
\cline{3-8}
\textbf{}                                  & \textbf{}                                                        & \multicolumn{2}{c|}{\textbf{Train}} & \multicolumn{2}{c|}{\textbf{Validation}} & \multicolumn{2}{c|}{\textbf{Test}} \\ \hline
\multicolumn{1}{|c|}{\textbf{Model}}       & \textbf{\begin{tabular}[c]{@{}c@{}}Removed\\  Step\end{tabular}} & \textbf{Accuracy}  & \textbf{F1}    & \textbf{Accuracy}    & \textbf{F1}       & \textbf{Accuracy} & \textbf{F1}    \\ \hline
\multicolumn{1}{|c|}{\multirow{2}{*}{kNN}} & Convolution                                                      & -         & -              & 0.9840      & 0.97/0.98/0.99    & 0.9888   & 0.98/0.98/0.99 \\
\multicolumn{1}{|c|}{}                     & Shift                                                            & -                  & -              & 0.8400               & 0.82/0.85/0.84    & 0.6519            & 0.70/0.58/0.67 \\ \hline
\multicolumn{1}{|c|}{\multirow{2}{*}{MLP}} & Convolution                                                      & 0.9966             & 0.99/0.99/0.99 & 0.9857               & 0.98/0.98/0.99    & 0.9884            & 0.98/0.98/0.99 \\
\multicolumn{1}{|c|}{}                     & Shift                                                            & 0.5919             & 0.63/0.33/0.75 & 0.5780               & 0.64/0.27/0.73    & 0.6078            & 0.70/0.39/0.67 \\ \hline
\end{tabular}
\end{center}
\caption{Ablation test on Geometric Shapes.}
\label{table:ablation_shapes}
\end{table*}


\begin{table*}[]
\begin{center}{}
\begin{tabular}{cc|cc|cc|cc|}
\cline{3-8}
\textbf{}                                  & \textbf{}                                                        & \multicolumn{2}{c|}{\textbf{Train}} & \multicolumn{2}{c|}{\textbf{Validation}} & \multicolumn{2}{c|}{\textbf{Test}} \\ \hline
\multicolumn{1}{|c|}{\textbf{Model}}       & \textbf{\begin{tabular}[c]{@{}c@{}}Removed\\  Step\end{tabular}} & \textbf{Accuracy}   & \textbf{F1}   & \textbf{Accuracy}      & \textbf{F1}     & \textbf{Accuracy}   & \textbf{F1}  \\ \hline
\multicolumn{1}{|c|}{\multirow{2}{*}{kNN}} & Convolution                                                      & 0.9048              & 0.91/0.88     & 0.9053                 & 0.91/0.88       & 0.9120              & 0.92/0.89    \\
\multicolumn{1}{|c|}{}                     & Shift                                                            & 0.8770              & 0.89/0.85     & 0.8806                 & 0.89/0.85       & 0.8850              & 0.90/0.86    \\ \hline
\multicolumn{1}{|c|}{\multirow{2}{*}{MLP}} & Convolution                                                      & 0.7478              & 0.77/0.72     & 0.7606                 & 0.78/0.72       & 0.7490              & 0.77/0.71    \\
\multicolumn{1}{|c|}{}                     & Shift                                                            & 0.7411              & 0.78/0.66     & 0.7520                 & 0.79/0.68       & 0.7433              & 0.78/0.67    \\ \hline
\end{tabular}
\end{center}
\caption{Ablation test on Aerial Images.}
\label{table:ablation_aerial_images}
\end{table*}

\subsection{Ablation Study}

This section describes a couple of changes to the Laguerre-Gauss preprocessing procedure
in order to better understand how each component affects the process of reducing the
feature space. Even though the procedure contains more than two steps, the steps tested
here are the ones that are theoretically allowed, since direct modifications of the
filter, or with it, would not make sense.

Tables \ref{table:ablation_shapes} and \ref{table:ablation_aerial_images} show the
accuracy and F1 scores after removing the convolution and shift step during 
preprocessing, respectively.

\subsubsection{Convolution}

This step consists of ignoring the convolution of the Fourier transform of the image
with the Fourier transform of the filter. Instead, the Fourier transform of the image
is calculated and the procedure remains unmodified, but uses this transformed representation.

From Table \ref{table:ablation_shapes} it seems that removing this step is actually good. This could be an effect of the lack of background noise in this images, where the object
to classify is not hidden by others. But on Table \ref{table:ablation_aerial_images} the 
performance is degraded on MLP. Also, note that while training, the model quickly reaches
the reported accuracy and the loss stops decreasing, meaning that the network has stopped
learning and started over-fitting.

\subsubsection{Unshifted Components}

Here, once the image is convolved the line profiles are directly calculated assuming
that the center of the image is the axis' origin. Basically, the convolution is not
rearranged and the frequency components are disorganized across the matrix.

From Table \ref{table:ablation_shapes} it is possible to see that removing this
step greatly degrades geometric shape classification performance. On Table \ref{table:ablation_aerial_images} the performance is also diminished. The main
differences are reported by the F1 score.

\section{Discussion}
\label{sec:discussion}

The introduced methodology can learn a robust classifier for Aerial Images and Geometric shapes
(see Tables \ref{table:geom_shapes_results} and \ref{table:aerial_images_results}).
Particularly for the Aerial Images, the model faces many challenges given the diverse set of
characteristics present in the images. 

As shown in \cite{attenuation_paniagua} the LGSF can avoid noise in the image by
ignoring the low- and high- frequencies. Furthermore, once the Fourier transform
of an image is computed, the random noise (usually white noise) behaves as a
constant in the Fourier frequency domain \cite{Smith_1976}. Also as seen in
Section \ref{sec:hilbert_space} the transform is linear so the ``data transform
is the sum of the signal and noise transforms" \cite{Smith_1976}. This feature might give robustness to adversarial attacks, but was not tested in this paper.

As stated in Section \ref{sec:methodology}, the parameter $\omega$ of the LGSF was fixed to
0.9, as a trade-off between prioritizing high frequencies (sharper edges) but being
conscious of the importance of lower frequencies, specially in images taken at high-altitudes,
where the objects in the surface are difficult to detect. The selection of this hyper-parameter
according to the characteristics of each data set (through hyper-parameter tuning) should return
better models. Specifically, for the Aerial Images, upon inspection of the
misclassified instances (in train, validation and test data sets), it was seen that most
errors occurred on images taken from high-altitudes, meaning that the classifier is constrained
to certain altitudes.

Additionally, since this methodology extracts features
of the images from the frequency domain it can be expected that it can generalize to diverse environments,
since it does not need to know its particularities (i.e., background noise: deserts, ocean, etc.), 
and instead can look for relevant shapes. In \cite{Sierra_Sosa_2016} they hypothesize a
similar result on emotion recognition from different languages. Overall, this will
mean a more robust classifier.

Figures \ref{fig:lp_shapes} and \ref{fig:lp_aerial} show the average line profile for
the training data. From them it is clear that they differ the most around the center of
the spectrum. Since both data sets were resized to images of size $64 \times 64$ the spectral
resolution is truncated. Determining an adequate spectral resolution (higher or lower) means
that a better model could be obtained; in the case of Aerial Images, from the average line
profiles, it can be seen that the line profiles could be truncated to obtain an even smaller
feature space and still have differentiating characteristics. This results particularly interesting for classification
tasks with many labels and variance within elements of the same class. For instance, Smith and Gray \cite{Smith_1976} say that line profiles differ the most at larger frequencies, where noise
is even smaller. The latter confirms that images of higher resolution might return a better
feature space.

Following on the topic of line profiles, their value lies in the fact that for different classes
it is possible to obtain peaks and valley of varying amplitudes and at different frequencies.
Since the synthesized profiles obtained in this work always lied in the axis, it remains
necessary to study how sampling profiles from other paths within the spectrum might help
to obtain a richer feature space.

Regarding the ablation study, the convolution step turned out to be relevant for classification
of Aerial Images. When tested on Geometric Shapes, its effect is counterproductive. This is
an important fact, given that it supports the LGSF as a good edge enhancer (which was not needed on Geometric Shapes, since they already had their edges clearly defined). On the other side, 
the shifting phase turned out to be really important when constructing the reduced feature 
space. This confirms that specific edge directions describe most of the images and a
feature space where every time different directions are sampled does not provide value.

An application where this methodology can be successfully applied is finding regions of interest
within large areas/images. Due to the simple models, inference on partitions of the image will quickly
return those places where a more expensive model should focus, avoiding the costs of running
it over the whole space.

\section{Conclusion}
\label{sec:conclusion}
This work introduced Laguerre-Gauss Preprocessing. It was shown that it can be successfully applied to robustly learn to classify aerial images with simple models. As a result, the model’s size footprint is reduced, as well as the training and inference time. The LGSF enabled edge enhancement and reduced the low- and high-frequency noise. The LGSF distributes homogeneously and smoothly the intensity in the Fourier spectrum due to its isotropic feature. This resulted in characteristic frequencies that allowed learning special/relevant shapes within an image. In addition, The LGSF can enhance any small changes in the image according to small changes in frequency.

The parameter $\omega$ tends to one in order to preserve the spatial frequency distribution in the image and perform the bandpass filter component from the LGSF, but it is important to analyze and propose a methodology to select the appropriate value depending on each data set.

This feature selection technique achieves models with a lower complexity (measured in terms of the VC dimensionality) and introduces a tool to develop a robust, quick, simplified and low-cost solution which can be deployed in aerial vehicles.

Future work could focus on extending this method for classification tasks with more categories where the shapes have high variance between elements of the same class.


\clearpage
\balance
\setlength{\bibsep}{0.0pt}
{\small
\bibliographystyle{ieee_fullname}
\bibliography{egbib}
}


\end{document}